# The semantic mapping of words and co-words in contexts

Loet Leydesdorff & Kasper Welbers

Amsterdam School of Communication Research (ASCoR), University of Amsterdam

Kloveniersburgwal 48, 1012 CX Amsterdam, The Netherlands

**Abstract**

Meaning can be generated when information is related at a systemic level. Such a system can be an observer, but also a discourse, for example, operationalized as a set of documents. The measurement of semantics as similarity in patterns (correlations) and latent variables (factor analysis) has been enhanced by computer techniques and the use of statistics; for example, in "Latent Semantic Analysis." This communication provides an introduction, an example, pointers to relevant software, and summarizes the choices that can be made by the analyst. Visualization ("semantic mapping") is thus made more accessible.

## Introduction

In response to the development of co-citation maps during the 1970s by Small (1973; Small & Griffith, 1974), Callon *et al.* (1983) proposed developing co-word maps as an alternative to the study of semantic relations in scientific and technology literatures (Callon *et al*., 1986; Leydesdorff, 1989). Ever since, these techniques for "co-word mapping" have been further developed, for example, into "Latent Semantic Analysis" (e.g., Landauer *et al*., 1998; Leydesdorff, 1997). These methods operate on a word-document matrix in which the documents can be considered as providing the cases (e.g., rows) to which the words are attributed as variables (columns).

Factor-analytic techniques allow for clustering the words in terms of the documents, or similarly, the documents in terms of the semantic structures of the words (*Q*-factor analysis). Singular value decomposition combines these two options, but is not so easily available in standard software packages such as SPSS. In this brief communication, we provide an overview and summary for scholars and students who wish to use these techniques as an instrument, for example, in content analysis (Danowski, 2009). A more extensive manual can be found at http://www.leydesdorff.net/indicators where the corresponding software is also made available. In this communication, we provide arguments for choices that were made when developing the software. Our aim is to keep the free software up-to-date, and to keep the applications as versatile and universally applicable as possible.

**The word-document matrix**

The basic matrix for the analysis represents the occurrences of words in documents. Documents are considered as the units of analysis. These documents can vary in size from large documents to single sentences, such as, for example, document titles. The documents contain words which can be organized into sentences, paragraphs, and sections. The semantic structures in the relations among the words can be very different at these various levels of aggregation (Leydesdorff, 1991, 1995). Thus, the researcher has first to decide what will be considered as relevant units of analysis.

Secondly, which words should be included in the analysis? An obvious candidate for the selection is frequency of word occurrences (after correction for stopwords). Salton & McGill (1983), however, suggested that the most frequently and least frequently occurring words can be less significant than words with a moderate frequency. For this purpose, these authors proposed a measure: the so-called "term frequency-inverse document frequency," that is, a weight which increases with the frequency of the term *i*, but decreases as the term occurs in more documents (*k*) in the set (of *n* documents). The tf-idf can be formalized as follows:

$$\text{Tf-Idf}_{ik} = \text{FREQ}_{ik} * [\log_2 (n / \text{DOCFREQ}_k)] \qquad (1)$$

The function assigns a high degree of importance to terms occurring more frequently in only a few documents of a collection, and is commonly used in information retrieval (Spark Jones, 1972). Given its background in practice, however, the measure has not been further developed into a statistics for distinguishing the relative significance of terms.

The proper statistics to compare the rows or columns of a matrix is provided by $\chi^2$ or—using the Latin alphabet—"chi-square" (e.g., Mogoutov *et al*., 2008; Van Atteveldt, 2005). Chi-square is defined as follows:

$$\chi^2 = \sum_i (Observed_i - Expected_i)^2 / Expected_i \qquad (2)$$

The chi-square is summed over the cells of the matrix by comparing for each cell the observed value with the expectation—calculated in terms of the margin totals of the matrix. The resulting sum values can then be tested against a standard table. Both the relevant routines and the chi-square table are now widely available on the Internet; for example, at http://people.ku.edu/~preacher/chisq/chisq.htm. (If the observed values are smaller than five, one should apply the so-called Yates correction; the corresponding statistics is available, for example, at http://www.fon.hum.uva.nl/Service/Statistics/EqualDistribX2.html.)

Our programs—to be discussed below in more detail and available from http://www.leydesdorff.net/indicators—provide the user with the chi-square values for each word as a variable (by summing the chi-square values for cells over a column of the matrix) and additionally a file "expected.dbf" which contains the expected values in the same format as the observed values in the file "matrix.dbf." The user can thus easily compute the chi-square values using Excel.[1] Furthermore, the comparison between observed and expected values allows for a third measure which is easy to understand, albeit not based on a statistics, namely, the value of observed over expected (*obs/exp*). As in the case of chi-square, *observed/expected* values can be computed for each cell. However, these values can also be summed, for example, over the columns in order to enable the analyst to assess to which extent a word (as a column variable) occurs above or below expectation.

In summary, one can use four criteria for selecting the list of words to be included in the analysis: (*i*) word frequency, (*ii*) the value of tf-idf, (*iii*) the contribution of the column to the chi-square of the matrix, and (*iv*) the margin totals of observed/expected for each word. In case studies, we found this last measure most convenient. However, all four measures are made available.

**The analysis**

The asymmetrical word-document matrix—in social network analysis also called a 2-mode matrix—can be transformed into a symmetrical co-occurrence matrix (1-mode) using matrix algebra. This can be done in both (orthogonal) directions, that is, in terms of co-words or co-occurring documents.[2] The resulting matrix is called an affiliations matrix in social network analysis, and is standardly available in software for social network analysis (such as Pajek and UCINet). This network is relational and allows, among other things, for the analysis of pathways. Pathways have been identified as indicators of inventions and innovations (e.g., Bailón-Moreno *et al.*, 2005 and 2007; Stegmann & Grohmann, 2003).

[1] Chi-square is not available for matrices in SPSS because SPSS presumes that the two variables have first to be cross-tabled.
[2] Technically, one multiplies the matrix (*A*) with its transposed as either $AA^T$ or $A^TA$.

The word-document matrix can also be analyzed in terms of its latent dimensions using factor analysis, multi-dimensional scaling (MDS) or singular value decomposition (SVD), etc. Note that factor analysis and SVD operate in the vector-space that is generated by first transforming the matrix using the Pearson correlation coefficients between the variables. In the vector space, however, similarity is no longer defined in terms of relations, but correlations among the distributions (vectors).

Since the distributions of words in texts are skewed (Ijiri & Simon, 1977), the use of the Pearson correlation—implying regression to the mean—is debatable (Ahlgren *et al.*, 2003). Salton's *cosine* has the advantage of not using the mean, but otherwise its formulation is completely analogous. Cosine-normalization of the variables therefore provides an attractive alternative, but one loses the advantage of orthogonal rotation possible with factor analysis and statistical testing (Bensman, 2003; White, 2003 and 2004). However, one can use the factor-analytic results to color the semantic maps based on cosine-normalized variables (Egghe & Leydesdorff, 2009). Note that it is preferable to factor analyze not the (1-mode) co-occurrence matrix, but the 2-mode word-document matrix if available as a result of the data collection (Leydesdorff & Vaughan, 2006).

In other words, the vector space can be approximated by constructing a semantic map on the basis of cosine-normalized variable patterns or by using factor analysis (or SVD), but these two representations will not be precisely similar because the factor analysis is based on the Pearson correlation matrix. Note that the transition to the vector space changes the perspective from a network perspective (as predominant in social network or co-word analysis) to a systemic perspective. The words are provided with meaning in terms of the semantic structures in the sets, and therefore one can legitimately use concepts such as "latent *semantic* analysis" and "*semantic* mapping."

The results can also be considered as a quantitative form of content analysis (Danowski, 2009; Carley & Kaufer, 1993; Leydesdorff & Hellsten, 2005). Unlike content analysis, however, the semantic structure (for example, the grouping of words as variables in terms of clusters or factors) is induced from the data and not provided on the basis of an *a priori* scheme. Thus, one can potentially reduce the so-called "indexer effect" (e.g., Law & Whittaker, 1992). Alternatively, one can use this technique for sets which are too large for qualitative content analysis or for the validation of content analysis using samples.

In summary, the development of statistical techniques has enabled us to move from Osgood *et al.*'s (1957) initial attempts to measure meaning using 7-point (Likert) scales to automated content analysis which provides us with semantic maps of the intrinsic meaning contained in document sets. Relevant software and techniques for these mapping efforts are available on the Internet.

## An empirical example

As an empirical example, we searched the Web-of-Science (WoS) of Thomson Reuters with the search string ‘ti= “impact factor” and py = (2008 or 2009)’ on November 12, 2010. This search resulted in 195 documents; these documents contain 59 words which occur more than twice (after correction for stop words; for example, at http://www.lextek.com/manuals/onix/stopwords1.html). Using the routine “ti.exe”—available at http://www.leydesdorff.net/software/ti/index.htm—one can perform factor analysis of the matrix and/or normalize the column variables using the cosine for the visualization. In Figure 1, the rotated factor matrix was used to colour the nodes in the map based on the cosine-normalized word occurrences over the documents.

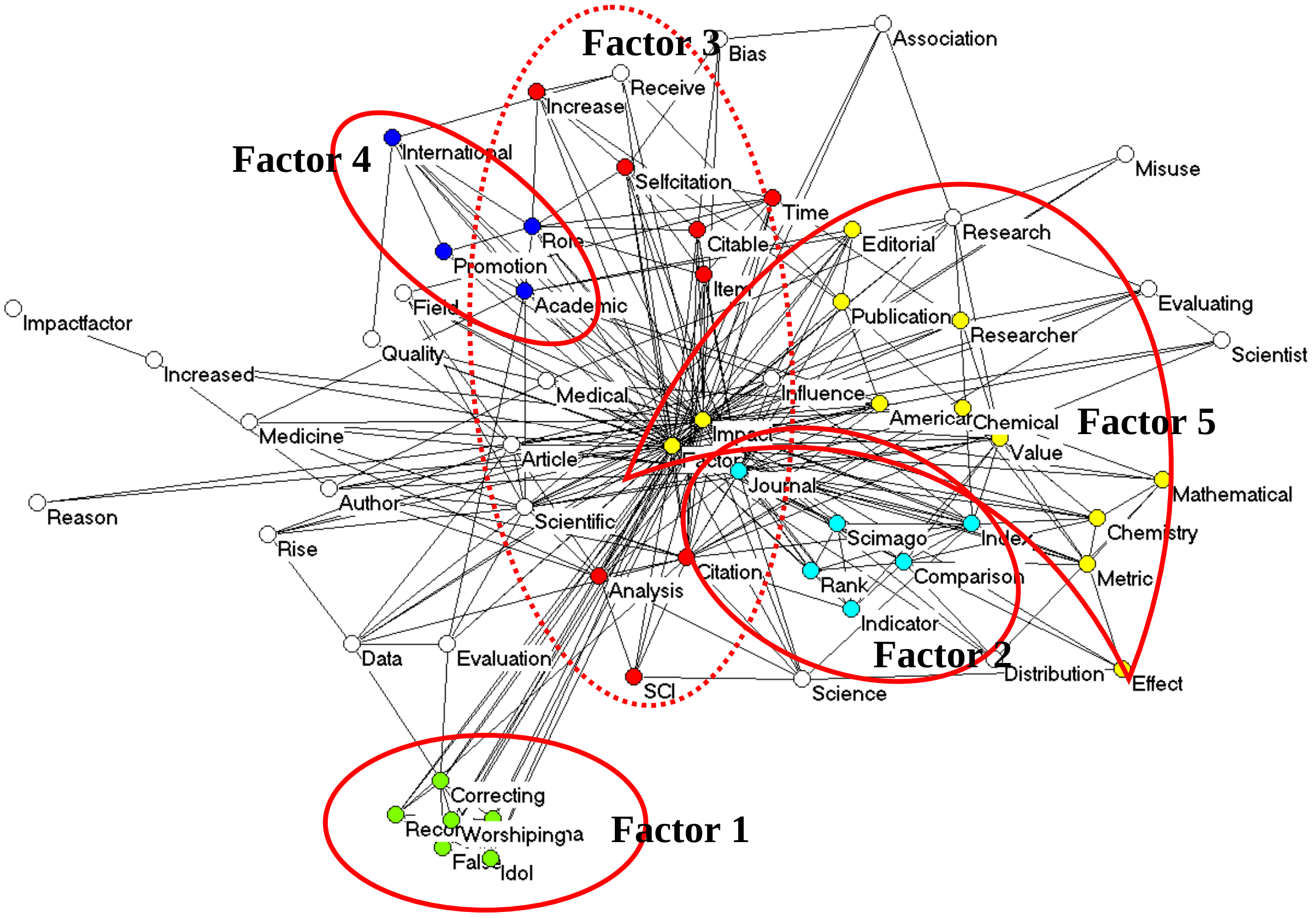

**Figure 1**: Five factors in 59 words as variables occurring more than twice in 195 documents with “impact factor” in the title and published in 2008 or 2009 (Kamada & Kawai, 1989; Factor loadings and cosine values < 0.1 are suppressed).

The results as shown in Figure 1 are not completely satisfactory; words which load on Factor 3 are displayed on both sides of the search terms "Impact" and "Factor". These two words which were also the original search strings are positioned in the center because they have a cosine similarity with all other words above the threshold level (cosine ≥ 0.1).

By changing to the observed/expected ratios, Figure 2 can be generated analogously. In this case, we do not use the word-document matrix with word frequency values, but the corresponding matrix of observed/expected ratios in each cell. Using this matrix as input to the factor analysis, the search terms "Impact" and "Factor" no longer load positively on any of the five factors, but exhibit interfactorial complexity. The (consequently different!) factor structure can in this case be penciled contingently on top of this semantic map. Note that many words do not exhibit factor loadings on any of the factors above the level of 0.1 (and are therefore left white).

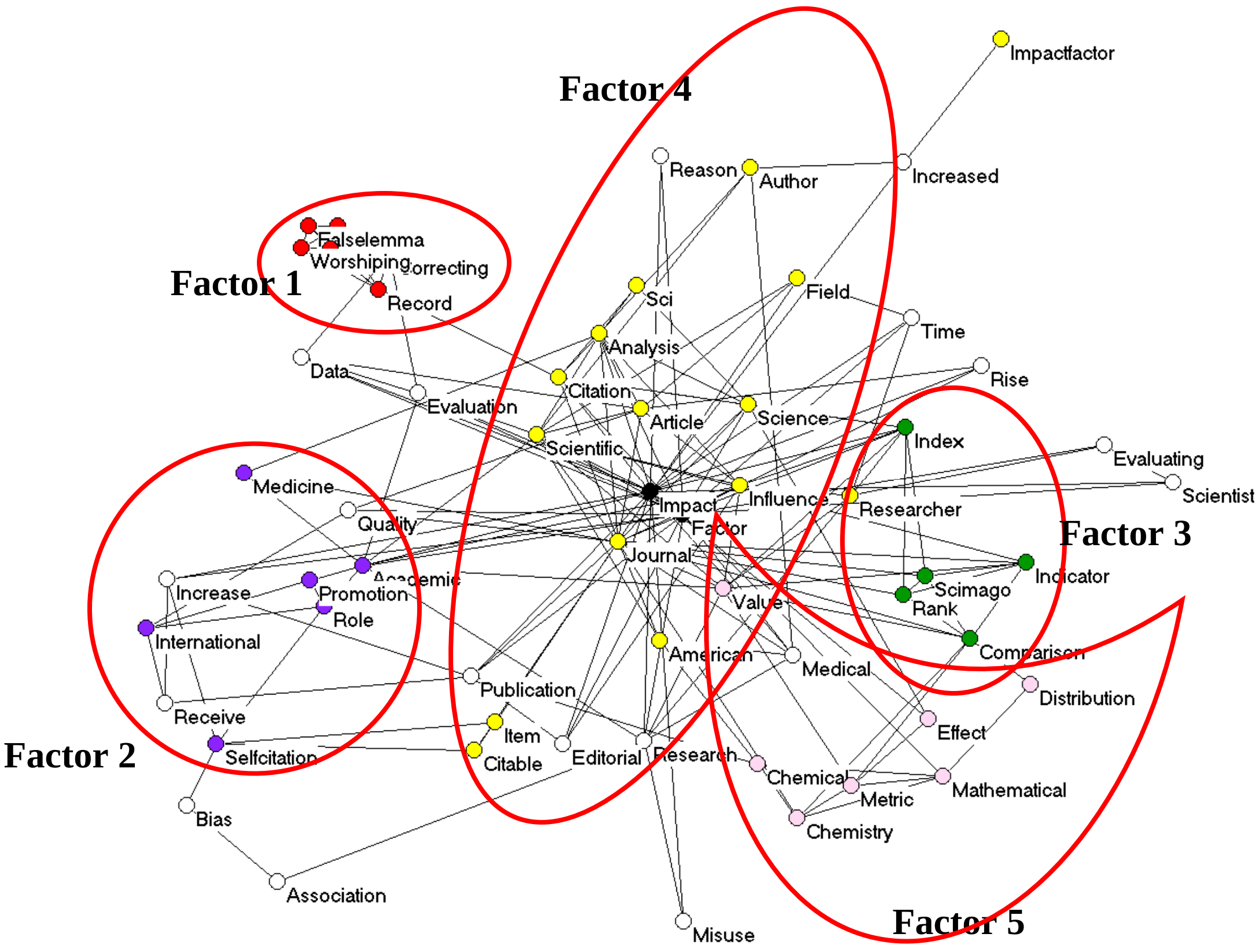

**Figure 2**: Five factors as in Figure 1, but now with observed/expected values as input instead of observed values.

In the above figures, only positive factor loadings were used for the coloring. Another visualization which includes also negative factor loadings can be generated by feeding the rotated component matrix directly into the visualization program as an asymmetrical (2-mode) matrix. This leads in this case to Figure 3. The factor loadings are by definition equal to the Pearson correlation coefficients among the variables (vectors) and latent dimensions (eigenvectors). The two constructs—vectors and eigenvectors—can thus be projected onto a single vector space (Leydesdorff & Probst, 2009).

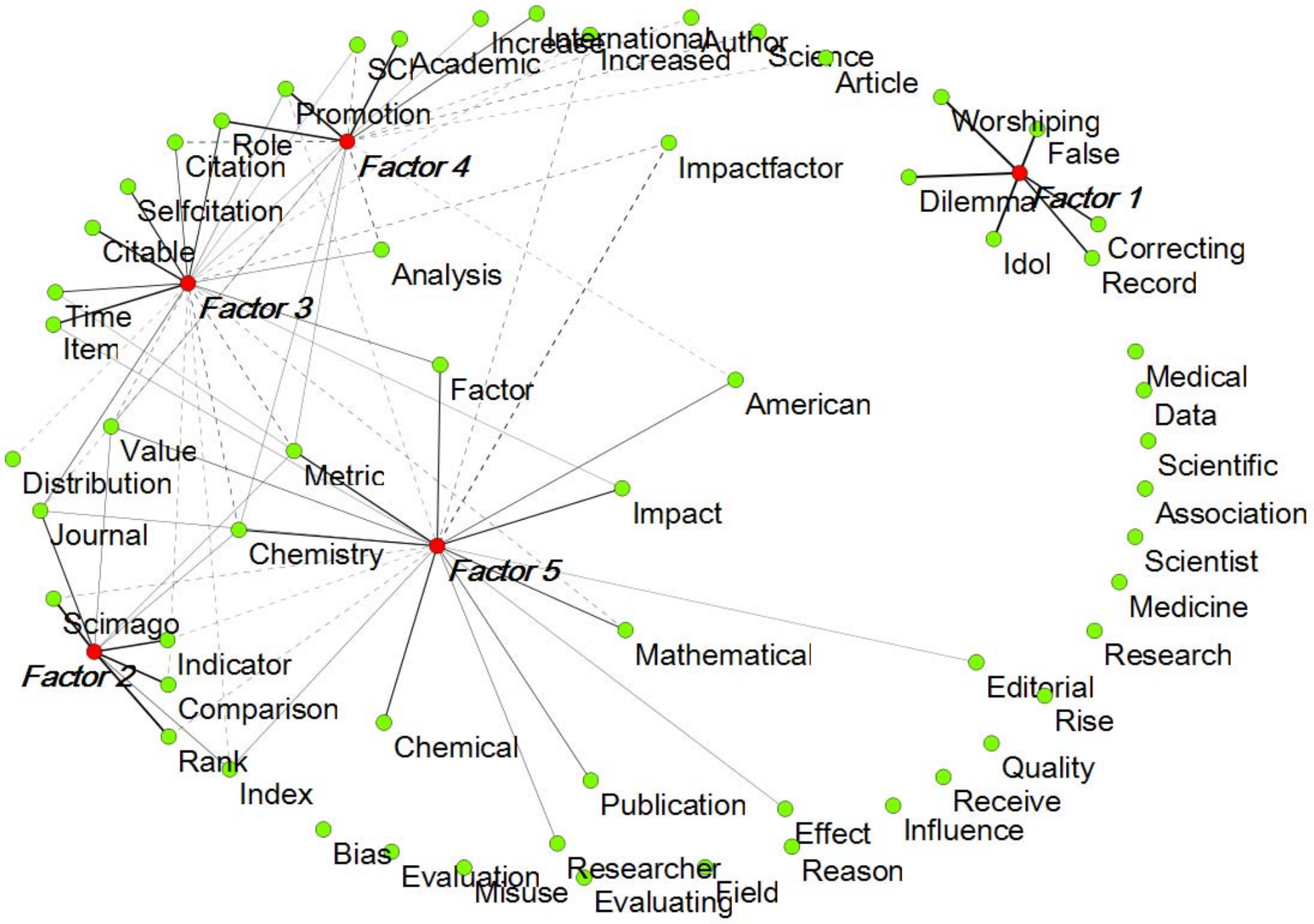

**Figure 3**: Visualization of the rotated factor matrix; dotted lines represent negative factor loadings (Fruchterman & Reingold, 1991). Factor loadings between -0.1 and +0.1 were suppressed.

One possible advantage of this representation is facilitation of the factor designation. For example, Factor 1 is otherwise isolated and indicates a set of words (a "frame"; cf. Hellsten *et al*., 2010; Scheufele, 1999) critical to the use of impact factors in research evaluation. Factor 5 seems most connected to the other three groupings; it shows the conceptual origins of "impact factor." This Factor 5 is most interrelated to Factor 3 which provides the frame of "citation analysis." The words "Factor," "Value," and "Metric" provide articulation points between these two star-shaped graphs. "Value" and "Metric" also relate to Factor 2, which indicates more recent ranking efforts.

Figures for different years can also be animated using a version of the network program *Visone* specially designed for this purpose (at http://www.leydesdorff.net/visone; Leydesdorff & Schank, 2008). In addition to statistical packages, the various output files (in the Pajek format) can also be imported into other network programs such as VosViewer available at http://www.vosviewer.org (Van Eck & Waltman, 2010) or the Social Networks Image Animator (SoNIA) at http://www.stanford.edu/group/sonia/ (Bender-de Moll & McFarland, 2006; Moody *et al*., 2005).

In addition to these possibilities for semantic mapping, the word-document matrix can, as noted, also be multiplied with its transposed in order to generate the word co-occurrence matrix. This matrix is made available by the programs in the Pajek format as the file coocc.dat. Figure 4 provides the resulting co-word map using all co-occurrences with a threshold of values larger than unity.

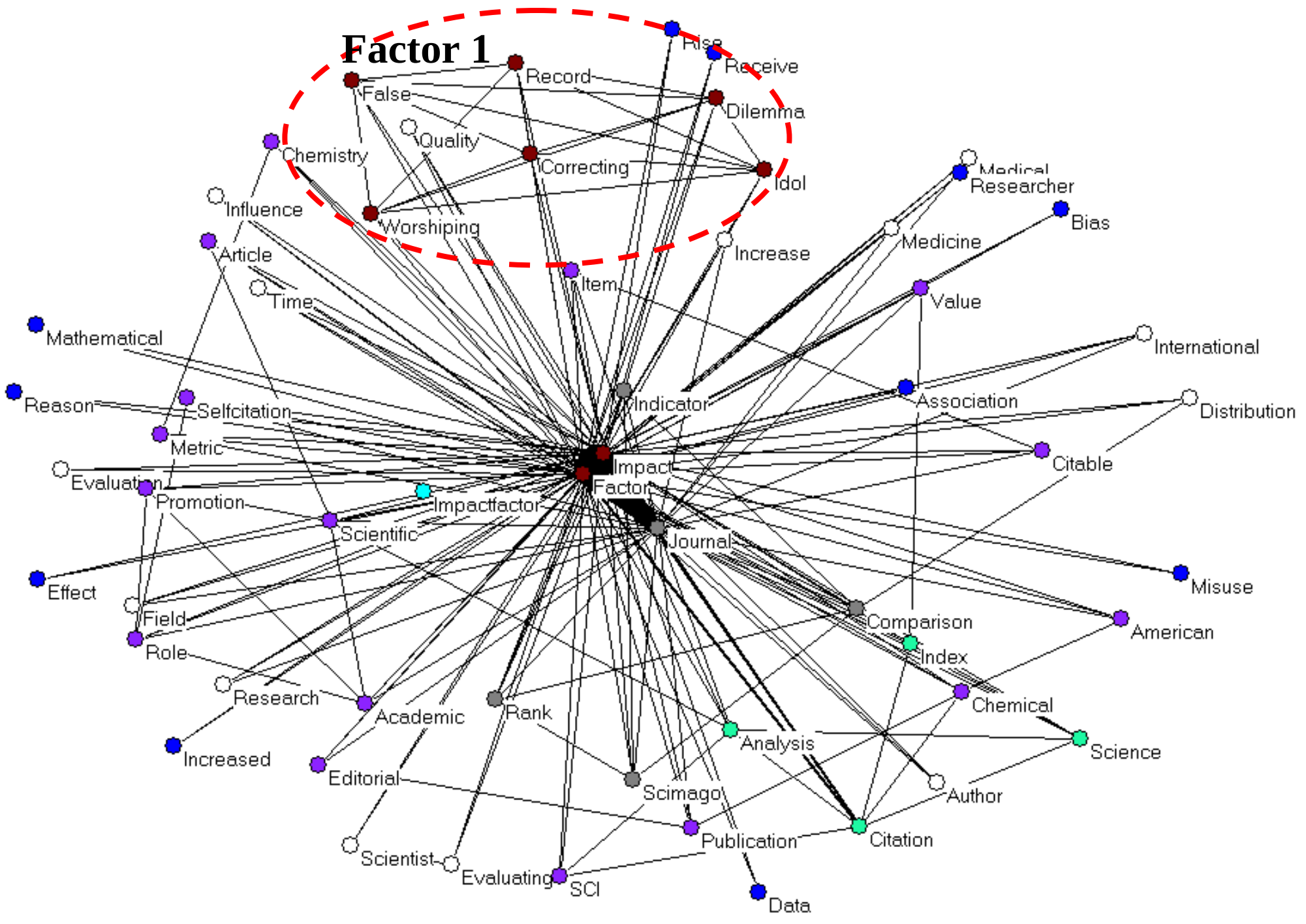

**Figure 4**: Co-word map of 59 words occurring more than twice in 195 documents with "impact factor" in the title and published in 2008 or 2008 (Kamada & Kawai, 1989; single co-occurrences of words are suppressed; *k*-core algorithm used for coloring the nodes).

In this map (Figure 4), the words "Factor" and "Impact" are very central because they co-occur in all items of the set. Other words with frequent co-occurrences such as "Journal" and "Indicator" occur in the vicinity of these two search terms. In this representation, Factor 1 distinguished above is still clearly visible at the top of the figure because these words—used exclusively in papers of critics of using impact factors—co-occur more in a grouping than others. Although the words attributed to Factor 2 can still be found as a separate group—colored in grey around the center—the distinction of structures upon visual inspection becomes increasingly difficult when one would attempt to retrieve factors which explain less of the common variance.

The star-shaped center-periphery structure dominates the graph in this network space and tends to overshadow the semantic structures of the vector space that provide the relations with meaning at the systems level. In the relational paradigm, however, relations are considered performatively, that is, as the potential construction of new meaning which crosses hitherto shaped divides and thus can function as indicators of invention and innovation (Bailón-Moreno *et al*., 2005 and 2007; Stegmann & Grohmann, 2003; cf. Leydesdorff, 1992).

**Conclusions and summary**

Despite the emphasis in the wording on semantics (as in "latent semantic analysis," "the semantic web" or "semantic mapping"), the measurement of the dynamics of meaning is still in its infancy. Meaning is generated when different bits of information are related at the systems level, and thus positioned in a vector space. Perhaps, one can define "knowledge" recursively as the positioning of different meanings in relation to one another.

Note that in the above, semantics is considered as a property of language, whereas meaning is often defined in terms of use (Wittgenstein, 1953), that is, at the level of agency. Ever since the exploration of intersubjective "meaning" in different philosophies (e.g., Husserl, 1929; Mead, 1932), the focus in the measurement of meaning has gradually shifted to the intrinsic meaning of textual elements in discourses and texts, that is, to a more objective and supra-individual level. The pragmatic aspects of meaning can be measured using Osgood *et al.*'s (1957) Likert-scales and by asking respondents. Modeling the dynamics of meaning requires another elaboration (cf. Leydesdorff, 2010).